\documentclass[letterpaper, 10 pt, conference]{ieeeconf}  

\IEEEoverridecommandlockouts                              

\usepackage{cite}
\usepackage{amsfonts}
\usepackage{amsmath}
\usepackage{amssymb}
\usepackage{graphicx}
\usepackage{textcomp}
\usepackage{xcolor}
\def\BibTeX{{\rm B\kern-.05em{\sc i\kern-.025em b}\kern-.08em
T\kern-.1667em\lower.7ex\hbox{E}\kern-.125emX}}
\usepackage{graphics} 
\usepackage{epsfig} 
\usepackage{mathptmx} 
\usepackage{times} 
\usepackage{amsmath} 
\usepackage{inputenc}
\usepackage[per-mode=symbol]{siunitx}
\usepackage{nicefrac}
\usepackage{xcolor}
\usepackage{eqparbox}
\usepackage{tabularx}
\usepackage{csquotes}
\usepackage{url}
\usepackage[caption=false,font=footnotesize]{subfig}
\usepackage{stmaryrd} 
\usepackage{makecell}
\usepackage{booktabs}
\usepackage[acronym, toc, shortcuts, nowarn]{glossaries}
\usepackage{multirow}

\usepackage{algorithm}
\usepackage{algpseudocode}

\definecolor{codegreen}{rgb}{0,0.6,0}
\definecolor{codegray}{rgb}{0.5,0.5,0.5}
\definecolor{codepurple}{rgb}{0.58,0,0.82}
\definecolor{backcolour}{rgb}{0.95,0.95,0.92}

\title{\LARGE \bf
    QBIT: Quality-Aware Cloud-Based Benchmarking for Robotic\\ Insertion Tasks
}

\author{Constantin Schempp$^{*, 1}$,  Yongzhou Zhang$^{*, 1, 2}$, Christian Friedrich$^{1}$, Björn Hein$^{1, 2}$
		\thanks{$^{1}$Karlsruhe University of Applied Sciences, 76133 Karlsruhe, Germany.
			{\tt\small \{name.surname\}@h-ka.de}}%
		\thanks{$^{2}$Karlsruhe Institute of Technology, 76131 Karlsruhe, Germany}
        \thanks{$^{*}$indicates equal contribution}
	}

\makeglossaries
\newacronym{CV}{CV}{Computer Vision}
\newacronym{LIDAR}{LIDAR}{Light Detection and Ranging}
\newacronym{AMR}{AMR}{Autonomous Mobile Robot}
\newacronym{IoT}{IoT}{Internet of Things}
\newacronym{ROS2}{ROS2}{Robot Operating System}
\newacronym{AMCL}{AMCL}{Adaptive Monte Carlo Localization}
\newacronym{SLAM}{SLAM}{Simultaneous Localization and Mapping}
\newacronym{DDS}{DDS}{Data Distribution Service}
\newacronym{RMSE}{RMSE}{Root Mean Square Error}
\newacronym{RTAB-MAP}{RTAB-MAP}{Real-Time Appearance-Based Mapping}
\newacronym{NSGA-II}{NSGA-II}{Nondominated Sorting Genetic Algorithm II}
\newacronym{PyGAD}{PyGAD}{Python Genetic Algorithm}
\newacronym{ICP}{ICP}{Iterative Closest Point}
\newacronym{MPPI}{MPPI}{Model Predictive Path Integral}
\newacronym{TRRT}{TRRT}{Transition-based Rapidly-exploring Random Trees}
\newacronym{DNaaS}{DNaaS}{Dex-Net as a Service}
\newacronym{RILaaS}{RILaaS}{Robot Inference and Learning as a Service}
\newacronym{VM}{VM}{Virtual Machine}
\newacronym{K8s}{K8s}{Kubernetes}
\newacronym{KCSS}{KCSS}{Kubernetes Container Scheduling Strategy}
\newacronym{KEIDS}{KEIDS}{Kubernetes-Based Energy and Interference Driven Scheduler}

\begin{document}

\maketitle
\thispagestyle{empty}
\pagestyle{empty}

\begin{abstract}

Insertion tasks are fundamental yet challenging for robots, particularly in autonomous operations, due to their continuous interaction with the environment.
AI-based approaches appear to be up to the challenge, but in production they must not only achieve high success rates. They must also ensure insertion quality and reliability.
To address this, we introduce QBIT, a quality-aware benchmarking framework that incorporates additional metrics such as force energy, force smoothness and completion time to provide a comprehensive assessment.
To ensure statistical significance and minimize the sim-to-real gap, we randomize contact parameters in the MuJoCo simulator, account for perceptual uncertainty, and conduct large-scale experiments on a Kubernetes-based infrastructure. Our microservice-oriented architecture ensures extensibility, broad applicability, and improved reproducibility. To facilitate seamless transitions to physical robotic testing, we use ROS2 with containerization to reduce integration barriers.
We evaluate QBIT using three insertion approaches: geometric-based, force-based, and learning-based, in both simulated and real-world environments. In simulation, we compare the accuracy of contact simulation using different mesh decomposition techniques. Our results demonstrate the effectiveness of QBIT in comparing different insertion approaches and accelerating the transition from laboratory to real-world applications. Code is available on GitHub \footnote[3]{\url{https://github.com/djumpstre/Qbit}}.

\end{abstract}

\section{INTRODUCTION}
Robotic insertion tasks, such as peg in hole, are fundamental in manufacturing and assembly processes \cite{Müller_Kutzbach_2019, Whitney_2004}. Despite advancements in AI and robotics, these tasks remain challenging due to continuous interaction between the robot and the environment and perception inaccuracies. While many innovative algorithms for robotic insertion tasks exist, they are often evaluated solely based on success rate as a single metric \cite{Zhang_Tomizuka_Li_2024}, overlooking crucial factors such as insertion quality, reproducibility, reliability, robustness, and alignment with industrial standards \cite{Marvel_Falco_2012}.

Benchmarks and data sets serve as critical driving forces in AI and robotics. 
Existing data sets and benchmarks, such as REASSEMBLE\cite{Sliwowski_Jadav_Stanovcic_Orbik_Heidersberger_Lee_2025} and RLBench\cite{james2020rlbench}, provide valuable simulated data or human-demonstrated trajectories. However, they primarily rely on the success rate as the sole metric, neglecting insertion quality. In contrast, industrial applications require the minimization of contact force to prevent surface damage, ensuring process repeatability, and improving reproducibility.

First, a comprehensive evaluation of the task-specific quality of insertion approaches is essential, considering industrial requirements such as minimizing maximum contact force to prevent surface damage.
Second, improving the accessibility and sharing of research findings is crucial to lowering entry barriers for new researchers and facilitating the reproduction and comparison of algorithms across various insertion tasks.
Third, the transition from simulation-based development and benchmarking to real-world deployment must be optimized to bridge the sim-to-real gap. Unlike tasks such as grasping, insertion tasks involve continuous interaction with the environment, making failures in real-world testing more likely to cause significant damage.

\begin{figure}[t]
	\vspace{2.5mm}
	\centering
    \includegraphics[width=0.46\textwidth]{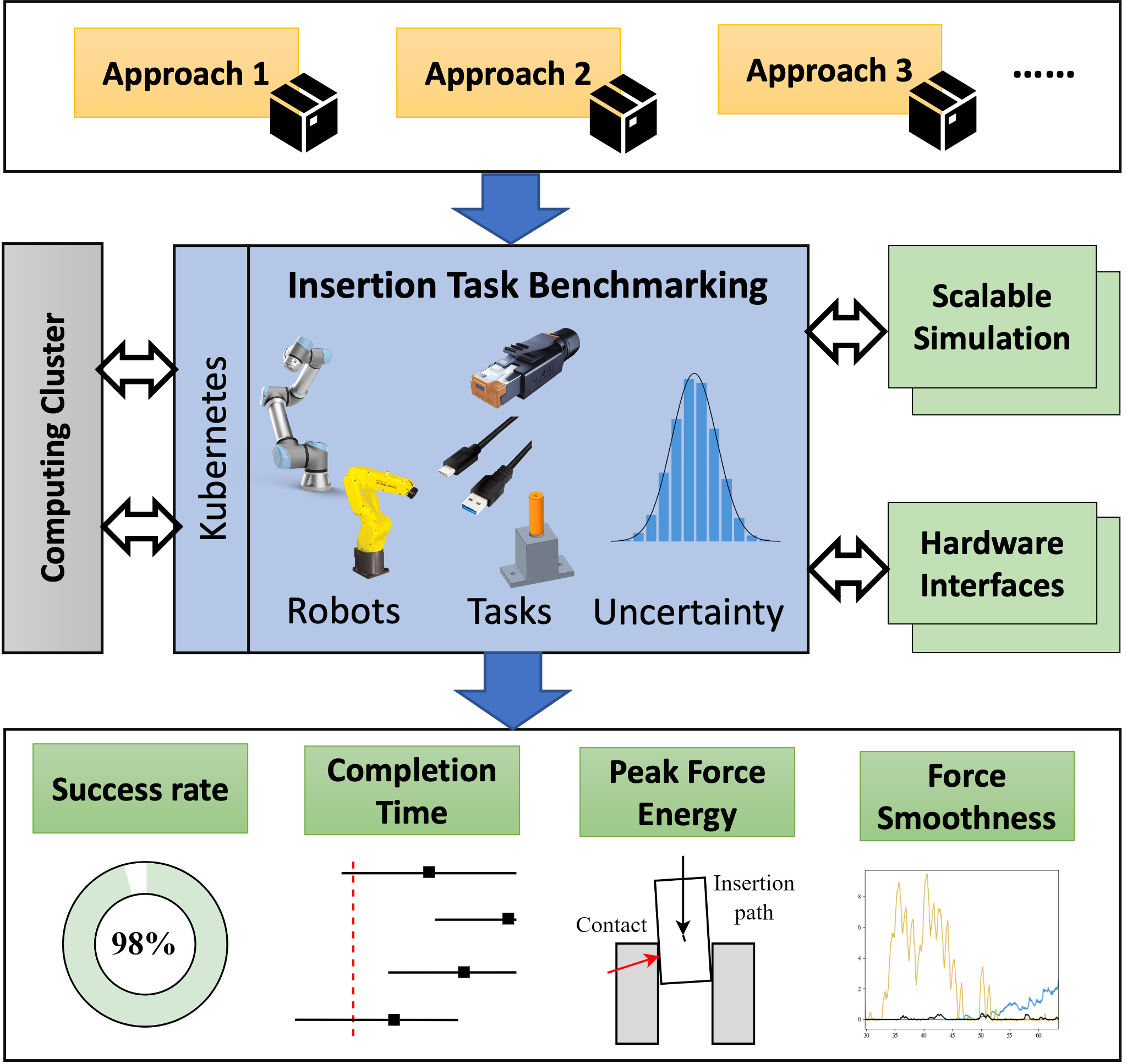}
	\caption{Overview of the QBIT benchmarking framework. It aims to use cloud computing and quality-aware metrics to evaluate and validate the insertion algorithms for with different task objects on different robots.}
	\label{fig:system_overview}
 \vspace{-0.5cm}
\end{figure}

To tackle these challenges, we introduce QBIT, a quality-aware, scalable, and extensible cloud-based benchmarking framework, see Fig.~\ref{fig:system_overview}.
Beyond success rate and task completion time, we incorporate two additional quality-related metrics: force energy and force smoothness.
Since camera calibration—both intrinsic and extrinsic parameters—as well as object localization errors in perception cannot be entirely eliminated, we model this uncertainty as a normal distribution. By conducting repeated tests, we ensure statistical significance in performance evaluation.

With extensibility and reproducibility in mind, we leverage cloud computing principles to develop the framework using a microservice-oriented architecture. All components, including algorithms, simulation environments, and hardware interfaces, are encapsulated within standalone containers. This design ensures easy integration of new robots and task objects, while also enabling seamless sharing within the research community to drive innovation in robotics.

The benchmarking framework supports both simulation-based and real-robot-based evaluation. For real-world validation, we employ ROS2 and ros2{\_}control as standard interfaces, facilitating reproducibility across different research groups and reducing implementation barriers.

Simulation-based benchmarking is often the preferred method for evaluating new algorithms. However, contact simulation in insertion tasks is inherently complex due to nonlinear behavior and parameter identification challenges.
To address this, we utilize the MuJoCo physics engine and conduct batch tests with randomized contact parameters, enhancing the likelihood of accurately modeling real-world contact force and torque distributions.

The combination of parameter randomization and perception uncertainty necessitates a large number of experimental runs. To expedite execution, we develop a lightweight experiment scheduler and controller, enabling tests to be efficiently conducted in a Kubernetes cluster, which can be self-hosted on edge servers or deployed via a public cloud provider.

To demonstrate its usability and effectiveness, we select three different approaches: a position-based, a force-control-based, and a learning-based approach for comparison of the proposed metrics in both simulation and real environment. 
Regular shapes, such as cylinders, and common industrial objects are used in the experiments.
We also conducted experiments to show how the proposed contact parameter randomization and the sphere-based decomposition can help to reduce the sim-to-real gap.

The main contributions of this paper are as follows:
\begin{itemize}
    \item QBIT Benchmarking Framework: A cloud-based, quality-aware benchmarking system for robotic insertion tasks.
    \item Enhanced Evaluation Metrics: Introduction of force energy and force smoothness for a more comprehensive assessment.
    \item Improved Contact Simulation: A sphere-based decomposition method for more accurate contact modeling in MuJoCo.
    \item Sim-to-Real Transfer: Randomized contact parameters and perception uncertainty modeling to improve real-world applicability.
    \item Scalable Cloud Execution: Kubernetes-based infrastructure enabling efficient large-scale benchmarking.
    \item Extensible Architecture: Modular microservice design for seamless integration of new robots and insertion approaches.
\end{itemize}

\section{RELATED WORK}
\label{sec:related_work}

Various approaches have been proposed to address the challenges associated with robotic insertion tasks.
Many learning-based approaches use reinforcement learning (RL) to guide the robotic insertion task \cite{Luo_Li_2021, Zang_Wang_Pan_Hou_Ding_Zhao_2024, Shi_Yuan_Tsitos_Cong_Hadjar_Chen_Zhang_2023}. In addition, imitation learning is used as an independent method or in conjunction with other approaches \cite{Wang_Beltran-Hernandez_Wan_Harada_2021, Ankile_Simeonov_Shenfeld_Agrawal_2024, Wang_Su_Sun_Chen_Xie_2024}. 
Furthermore, classical approaches use a combination of searching, force-control and geometric prior knowledge \cite{Wang_Liu_Liu_Huang_Yang_2024, Mei_Li_Tan_Zhu_Li_Zhou_2024}. Despite or perhaps because of these advancements, reproducibility and benchmarking remain significant challenges due to the lack of standardized evaluation criteria beyond success rates. Most existing benchmarks assess insertion tasks based only on success rate and completion time, overlooking critical quality factors such as force exertion and smoothness.

Contact simulation plays a crucial role in benchmarking insertion tasks but remains difficult due to nonlinear contact interactions. The finite element method (FEM)\cite{Bielak_2024} and methods that consider the surface microstructure require massive computation time and are therefore not fully suitable for robotics in general. The widely used robotics frameworks such as Pybullet\cite{coumans2020}, Gazebo\cite{koenig2004gazebo}, and MuJoCo\cite{todorov2012mujoco} are developed with a focus on real-time performance. Numerous studies are conducted to compare the accuracy and performance \cite{erez2015simulation, acosta2022validating}.
Among them, MuJoCo, which we used in this work, provides the most detailed modeling approach to simulate the contact interaction\cite{todorov2011contact, todorov2014contact}. However, in the end all of these frameworks require extensive manual parameter tuning for accurate contact modeling, which is both time-consuming and impractical for large-scale benchmarking.

In the context of cloud-based benchmark, the cloud-based competition and benchmark ORCTOC \cite{liu2021ocrtoc} for grasping and manipulation provided both simulation environments and remote access through cloud to the real robot setup for competition purposes. Compared to that, we build the benchmark with the microservice-oriented architecture to provide an interface for integrating new robots, and task objects and utilize the cloud computing to run the scalable experiments to get a statistical significance.

Regarding cloud-based development and testing, AWS RoboMaker provides a cloud service to build and run the simulation built in Gazebo at scale \cite{aws2025robomaker}. The Experiment Management System (EMS{\textregistered}), focuses on deploying the jobs to the public clouds to support the data-driven robotics experiments \cite{lin2023ems}. FogROS2 provides a secure communication middleware to offload the robotics software to the cloud with automatic resource provisioning to develop and deploy new robotics applications \cite{ichnowski2023fogros2}.
Moreover, in previous work, we proposed BatchJob \cite{zhang2024conquering} with KubeROS \cite{zhang2023kuberos} for simulation-based testing, which is mainly for the system level with a large number of containers and existing benchmark datasets. The start and stop mechanisms in the container scheduling introduced too much overhead for the scaling of this benchmark framework. Therefore, we propose a lightweight framework and scheduling mechanism for single-task benchmarking. The idea of direct execution in Kubernetes cluster aims to improve the usability by reducing the software dependencies.

\section{Quality-Aware Metrics}
\label{sec:quality_aware_metrics}
Traditional evaluation metrics for robotic insertion tasks, such as success rate and completion time, provide limited insights into the overall quality of the task execution, especially for many industrial use cases. We propose additional metrics that take into account the forces incurred during the entire insertion process.
These metrics are quality-related, since different insertion approaches show the same high success rate but result in large differences in contact forces, which is crucial for industrial assembly tasks.
Fig.~\ref{fig:sketch_insertion_task} shows the insertion task with the proposed metrics. We postulate that the following metrics will help provide a more comprehensive and realistic evaluation of robotic insertion strategies, bridging the gap between laboratory testing and industrial deployment.

\subsection{Insertion Task Formulation}
\begin{figure}
    \vspace{2mm}
    \centering
    \includegraphics[width=\linewidth]{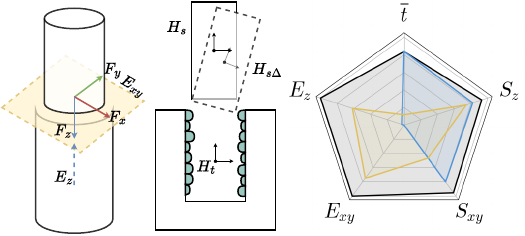}
    \caption{Insertion task formulation. Left: Yellow plane represents force signal energy metric $E_{xy}$ orthogonal to the insertion direction $E_z$. Center: Start pose $H_s$ and target pose $H_t$ and hole with surface roughness. Right: Resulting metrics of different approaches solving insertion task on a real robot.}
    \label{fig:sketch_insertion_task}
\end{figure}

The insertion task begins from an initial pose, denoted as: $H_s\in\mathbb{R}^6$,
which includes both position $(x, y, z)$ and orientation $(\alpha, \beta, \gamma)$, see Fig.~\ref{fig:sketch_insertion_task}.
During the insertion, we measure the contact wrench at the robot's end effector,

\begin{equation}
    \mathbf{w} \;=\; 
    \begin{bmatrix}
        \mathbf{F}, \mathbf{T} \\
    \end{bmatrix}^T,
\end{equation}
where \(\mathbf{F} \in \mathbb{R}^3\) represents the measured contact force and \(\mathbf{T} \in \mathbb{R}^3\) the measured contact torque.
To ensure statistical significance, each insertion trial is repeated K times. 
Hence, for each trial \(i \in \{1, 2, \ldots, K\}\), the discrete \emph{time-series} of measured wrenches is:
\[
\mathbf{w}_i(n) 
\;=\; 
\begin{bmatrix}
\mathbf{F}_i(n), \mathbf{T}_i(n)
\end{bmatrix}^T, \quad
n \in [0, N].
\]
Since torque is a linear mapping of the incurred forces, the proposed quality metrics are based only on force to reduce redundancy.

\subsection{Statistic with Uncertainty}
To account for uncertainties in camera calibration, object localization, and perception errors, we model the initial insertion pose as a normally distributed variable:
\begin{equation}
    H_{s\Delta} \sim \mathcal{N}(H_s,\,\Sigma_H),
\end{equation}
where 
\(\Sigma_H\) represents the respective covariance.

Pose accuracy varies between robotic systems. It is higher for industrial and lower for service robots.
Hence, it is necessary to consider the accuracy in the benchmark environment. This can be achieved by
sampling a position error
from a normal distribution with zero mean and the repeatability error of the robotic system as specified by the manufacturer as standard deviation and adding it to the desired target position during execution.
Additionally, to simulate sensor noise, we introduce Gaussian white noise into the force-torque sensor data recorded in MuJoCo. By performing and recording multiple test repetitions \(\{\mathbf{w}_1(n),\, \mathbf{w}_2(n),\, \dots,\, \mathbf{w}_K(n)\}\) under these varying conditions, we ensure a statistically robust evaluation of insertion performance and strategy effectiveness.

\subsection{Force Energy}
Minimizing force exertion during insertion is crucial for ensuring smooth assembly and preventing surface damage\cite{Marvel_Falco_2012}. 
To quantify this, we define force energy $E$ as the cumulative force applied over the course of the insertion process:
\begin{equation}
    E = \frac{1}{N} \sum_{n=0}^{N} |\mathbf{F(n)}|^2,
\end{equation}
where  $\mathbf{F} = \mathbf{F_z}$ to calculate the energy $E_z$ in the direction of insertion and $\mathbf{F} = \mathbf{||F_x,F_y||}$ to calculate the energy $E_{xy}$ of the incurred forces on the plane orthogonal to the insertion direction, see Fig. \ref{fig:sketch_insertion_task}.
Lower $E$ values indicate a more controlled and efficient insertion process, making this metric a key indicator of insertion quality.

\subsection{Force Smoothness}
A stable insertion process requires minimizing abrupt fluctuations in applied force. To quantify this, we define force smoothness $S$ as the standard deviation of the force signal’s rate of change:
\begin{equation}
    S = \sqrt{ \frac{\sum(\mathbf{\dot{F}} - \mu)^2}{N}}.
\end{equation}
As before, we distinguish between the direction of insertion $\mathbf{\dot{F}} = \mathbf{\dot{F}_z}$ and the plane orthogonal to it
$\mathbf{\dot{F}} = \mathbf{||\dot{F}_x,\dot{F}_y||}$. To mitigate sensor noise, force signals in real-world experiments are filtered using a low-pass filter before computing smoothness.
\subsection{Success Rate and Completion Time}
In addition to the new quality-related metrics presented here, we use the success rate $R$ and mean completion time $\overline{t}$ as complementary metrics:
\begin{align}
    R = \frac{1}{K} \sum_{i=1}^{K} b_{i} \in [0, 1] ,\quad \overline{t} &= \frac{1}{K} \sum_{i=1}^K t_i,
\end{align}
where $b_{i}$ is a binary value determining if repetition $i$ of the task was successful, $t_i$ the time required to complete repetition $i$ and $K$ the total number of repetitions.

\section{Framework Architecture}
\label{sec:bm_system}
Our benchmarking framework is designed to be lightweight, modular, and scalable, ensuring ease of setup, usage, and extensibility. It enables seamless integration of new robots, task objects, and algorithms, making it accessible to both research and industry. It is able to utilize the computing resources of local servers and public cloud to accelerate simulation-based benchmarking with randomization, and its architecture supports both simulation-based and real-robot evaluations, streamlining the transition from virtual testing to real-world deployment.

\subsection{System Overview}

\begin{figure}
    \vspace{2mm}
    \centering
    \includegraphics[width=0.95\linewidth]{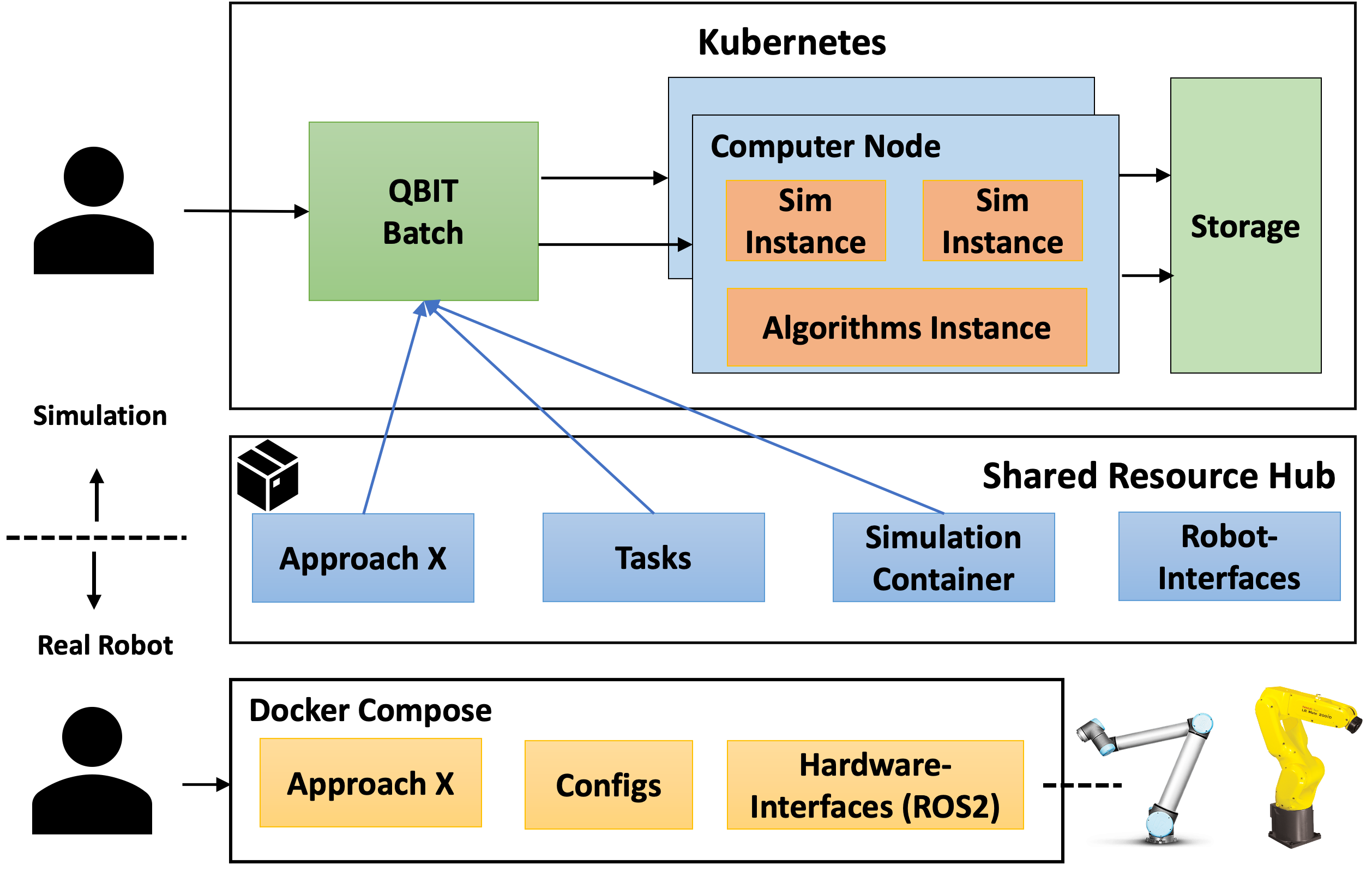}
    \caption{Overview of the benchmark software architecture. }
    \label{fig:qacbi_sw_overview}
\end{figure}

The framework follows a microservice-oriented architecture, where each component - ranging from insertion task algorithms and simulation environments to hardware interfaces - is containerized. This modular design enhances flexibility, scalability, and ease of integration for new robots, task objects, and insertion approaches, see Fig~\ref{fig:qacbi_sw_overview}. With this architecture, the components can be easily shared between research groups and adapted or extended depending on the available robots and tasks. The resource under this framework is shared in container level and uses the public container registry such as DockerHub.

For simulation-based benchmarking, the system leverages Kubernetes clusters, which can be deployed on self-hosted edge servers or public cloud providers. For the transition from simulation-based benchmarking to real robot-based validation, we containerize the hardware interface software for each robot and define the interface (joint position/torque, end effector position/velocity) same as for simulation.

\subsection{Scalable Simulation-based Benchmarking}
Simulation-based benchmarking can be easily integrated into the daily development workflow to evaluate the algorithms after each change. However, we cannot model the insertion process precisely due to the complex contact phenomena. It is also time-consuming to model and fine-tune for each task object. Therefore, we propose to randomize the contact parameter in the simulation to cover the real contact condition. In addition, since the uncertainty from perception to localize the target objects cannot be eliminated, we also take this as described in sec. \ref{sec:quality_aware_metrics} and generate the test runs to evaluate the reliability of the proposed approach.

In the current implementation, we use MuJoCo as the physics engine because it provides complex contact models with multiple parameters, such as contact shape (cone), stiffness, damping, and friction coefficients to simulate the different contact behaviors. It is also well optimized to simulate a large number of bodies with different numerical solvers, such as Newton, and Runge-Kutta-4, and thus suitable for scaling.

\begin{figure}
\vspace{2mm}
    \centering
    \includegraphics[width=0.95\linewidth]{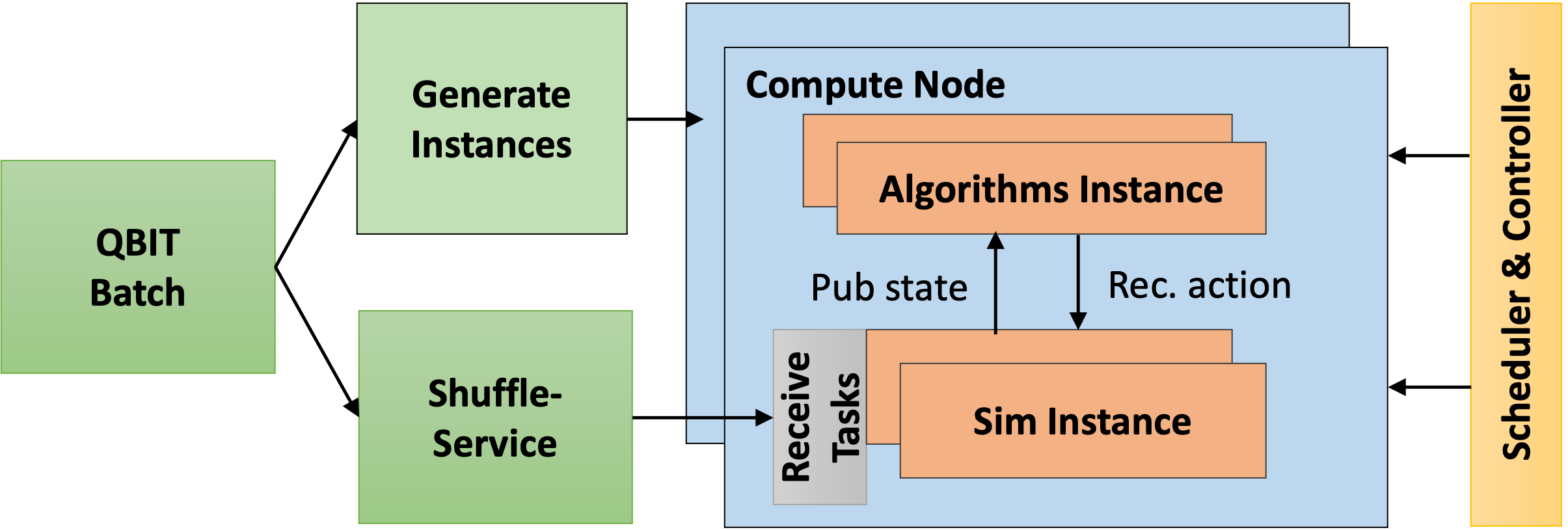}
    \caption{Asymmetrically scale the number of tested algorithm instances and simulation instances to speed up experiments.}
    \label{fig:qacbi_batch}
\end{figure}

To efficiently execute a large number of runs, we propose a lightweight batching approach to speed up the benchmarking, as shown in Fig.~\ref{fig:qacbi_batch}. We use a YAML file to specify the container image address of the tested algorithms, the simulation instances, the randomized simulation parameter, and the uncertainty for the starting pose. The QBIT batch then calls the Kubernetes API server to create the instances and uses a shuffle service to generate the task queues that are sent to each simulation instance. Due to the effect that the simulation runtime in a control loop is not equal to the execution time (inference time) of the insertion algorithm, especially when we use fine mesh to improve the contact simulation accuracy. We wrap the algorithm as a server and provide the services to the simulation instance that as client using gRPC \cite{grpc2025} to reduce the waiting time. For example, we can start 2 inference instances with 20 simulation instances to make maximum use of the resources in one compute node.

\subsection{Hardware Interface with Real Robots}

To validate the results obtained from simulation-based experiments, validation on real robots is still necessary. Unlike simulation, 
reproducing the result on hardware in different research groups is not easy. With our microservice-oriented architecture, only the container with the hardware interface needs to be adapted. Since ROS2 \cite{macenski2022robot} is now the de facto framework, we use ros2{\_}control and MoveIt2 \cite{Coleman2014ReducingTB} as the libraries to control the real robots. This component is also containerized. For the interface, we currently use joint position and end effector position/velocity control in both simulation and real robots, as it is mostly used in the current robot learning community. Direct torque control is suitable for insertion tasks, but only few robot manufacturers provide this low-level control interface. This framework is designed with high extensibility to integrate new robots with different interfaces.

\section{EXPERIMENTS}
To demonstrate our proposed benchmarking approach, we conducted experiments with both, simulated and physical robots. 
Furthermore, it is shown that different methods using different modalities can ultimately lead to successful insertion; however, they exhibit significant differences in terms of the forces incurred and the quality achieved.
\subsection{Setup}
\begin{figure}
    \vspace{2.3mm}
    \centering
    \includegraphics[width=0.95\linewidth]{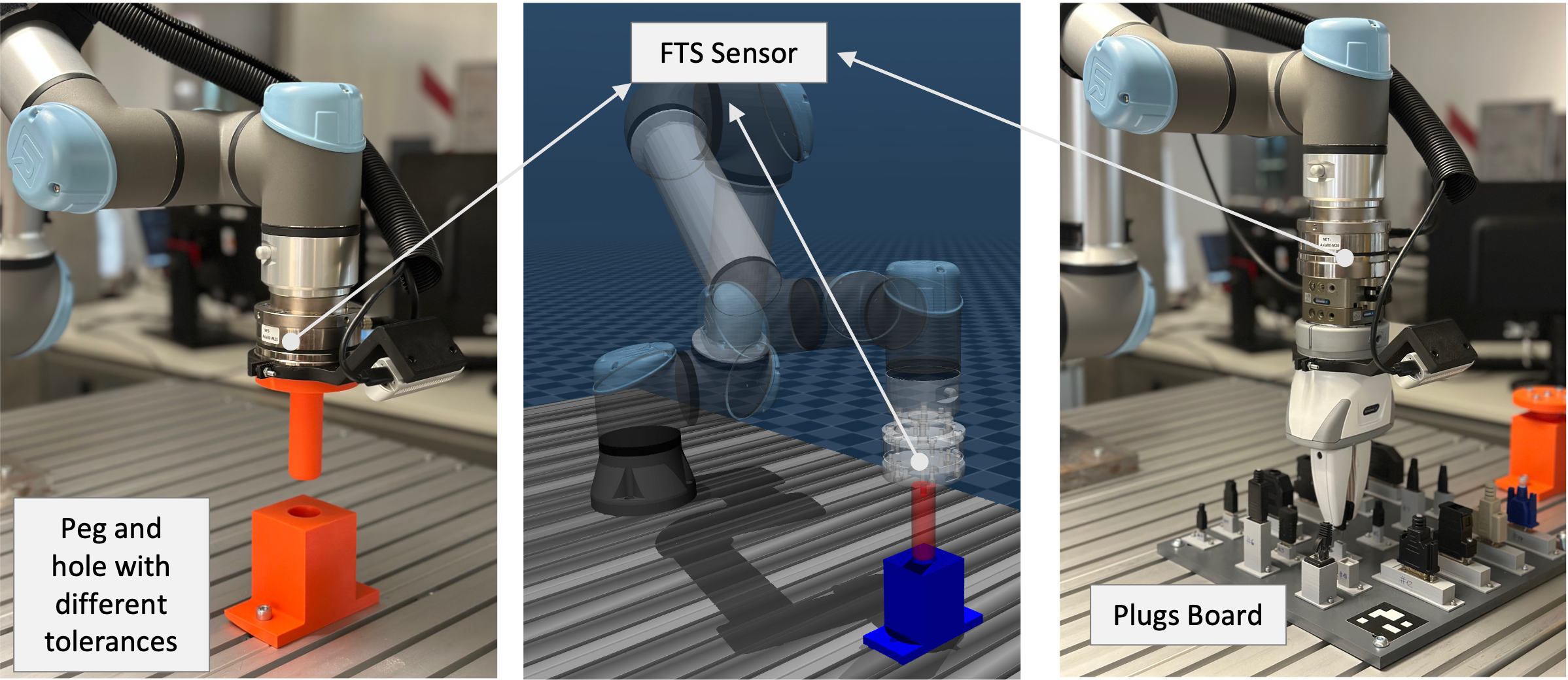}
    \caption{Experimental setup. Left: Real robot system with peg in hole task, Center: Simulation environment and Right: Real robot system with electrical connectors task. Interface is extensible to different robots and task objects.}
    \label{fig:experimental-setup}
\end{figure}
For the experiments on the real robotic system, we use an UR5e with a peg fixed to the robot end effector to remove the uncertainty of grasping. For measuring the force and torque during insertion, a six-axis force-torque sensor FTN-AXIA80 by the company SCHUNK is used.
Due to the fact that the simulation can be scaled more easily than real experiments, we are able to explore a multitude of insertion tasks and environmental properties. The simulated tasks consist of different peg in hole scenarios under high, $1$ mm, and low, $0.1$ mm, tolerance and small object insertion like USB plugs. Environmental properties are adapted in simulation, which regard for contact stiffness, damping and friction.
Three insertion approaches were implemented in both the real and simulated environment.
For each approach, we sample an uncertain initial pose $H_{s\Delta} \sim \mathcal{N}(H_s,\,\Sigma_H)$ above the hole and guide the robot using the respective policy. Experimental setup is shown in Fig. \ref{fig:experimental-setup}.

\subsection{Insertion Approaches}
\begin{figure}[!t]
    \centering
    \includegraphics[width=\linewidth]{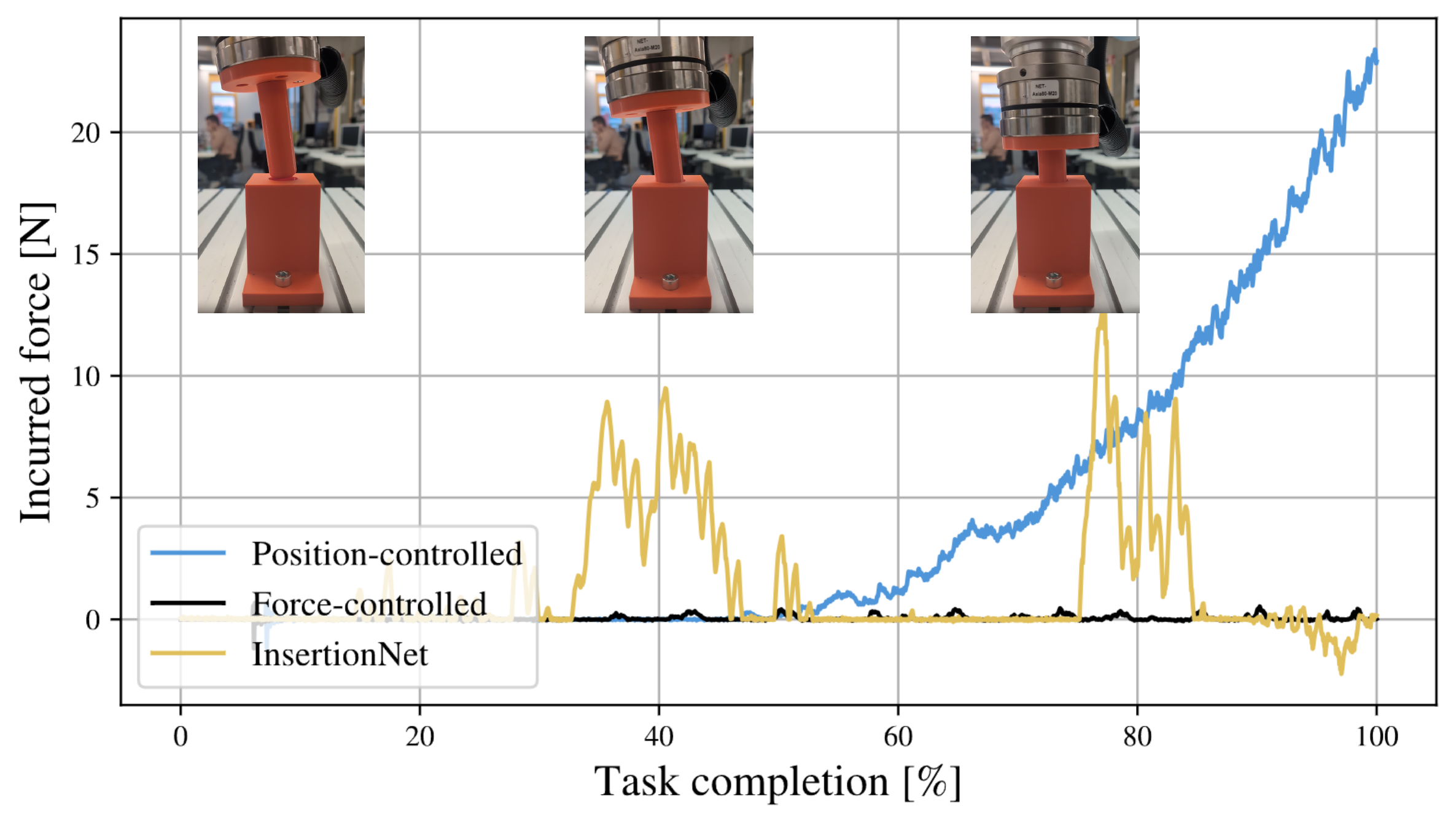}
    \caption{Peg in hole insertion task using UR5e. Each approach starts at same initial pose $H_{s\Delta}$. }
    \label{fig:approach-comparison}
    \vspace{-2mm}
\end{figure}
\textbf{\textit{Position-controlled:}} We implement position control, which is compatible with all robotic systems. This method does not account for external forces when guiding the robot's end-effector. It serves as a baseline for comparing simulated force calculations with real-world measurements, as it is highly controllable and influenced by fewer unknown factors. 

\textit{\textbf{Force-controlled:}} We implement the following classical control approach \cite{yoshikawa2000force} to guide the peg in hole process:
\begin{align}
    \Ddot{x}_t &= \frac{1}{M} (w_d - w_a - D \dot{x}_t - C x_t) \\
    \dot{x}_{t+1} &= \dot{x}_t + \Ddot{x}_t \cdot \Delta t,
\end{align}
where $w_a$, $w_d$ $\in \mathbb{R}^{6}$ are actual wrench and desired wrench and $M$, $D$, $C$ $\in \mathbb{R}^{6 x 6}$ the diagonal mass, damping and stiffness matrices of the control law. We set a desired force in $z$-direction to achieve downward movement. All other values in $w_d$ are set to zero.
The end effector twist $\dot{x}_{t+1}$ is used to guide the robot.

\textbf{\textit{Learning-based:}} We utilize InsertionNet (IN) \cite{Spector_Castro_2021} as a learning-based approach to solve the peg in hole task. 
It is a multi-modal approach fusing together force and vision information. Data is collected offline and used to train a neural network to predict the residual pose guiding the robot to the target.
As no publicly available codebase for the original approach was found, we developed our own implementation. 
The objective is to find a policy $\pi_{residual}$ that maps a tuple of image $\in \mathbb{R}^{H\times W\times C}$ and wrench $\in \mathbb{R}^6$ into action $a$. Formally
\begin{equation}
    \pi_{residual} \quad : \quad \mathbb{R}^{H\times W\times C} \times \mathbb{R}^6 \xrightarrow{} \mathbb{R}^6,
\end{equation}
where the action $a \in \mathbb{R}^6$ is the end effector pose correction in Cartesian space. 
InsertionNet's advantages are its generalization and flexibility. 
It exhibits robustness to socket pose and variance in plug color or shape.
For further information about training data collection and network architecture, the reader is referred to the original paper.

\begin{figure}
\vspace{2.4mm}
    \centering
    \includegraphics[width=\linewidth]{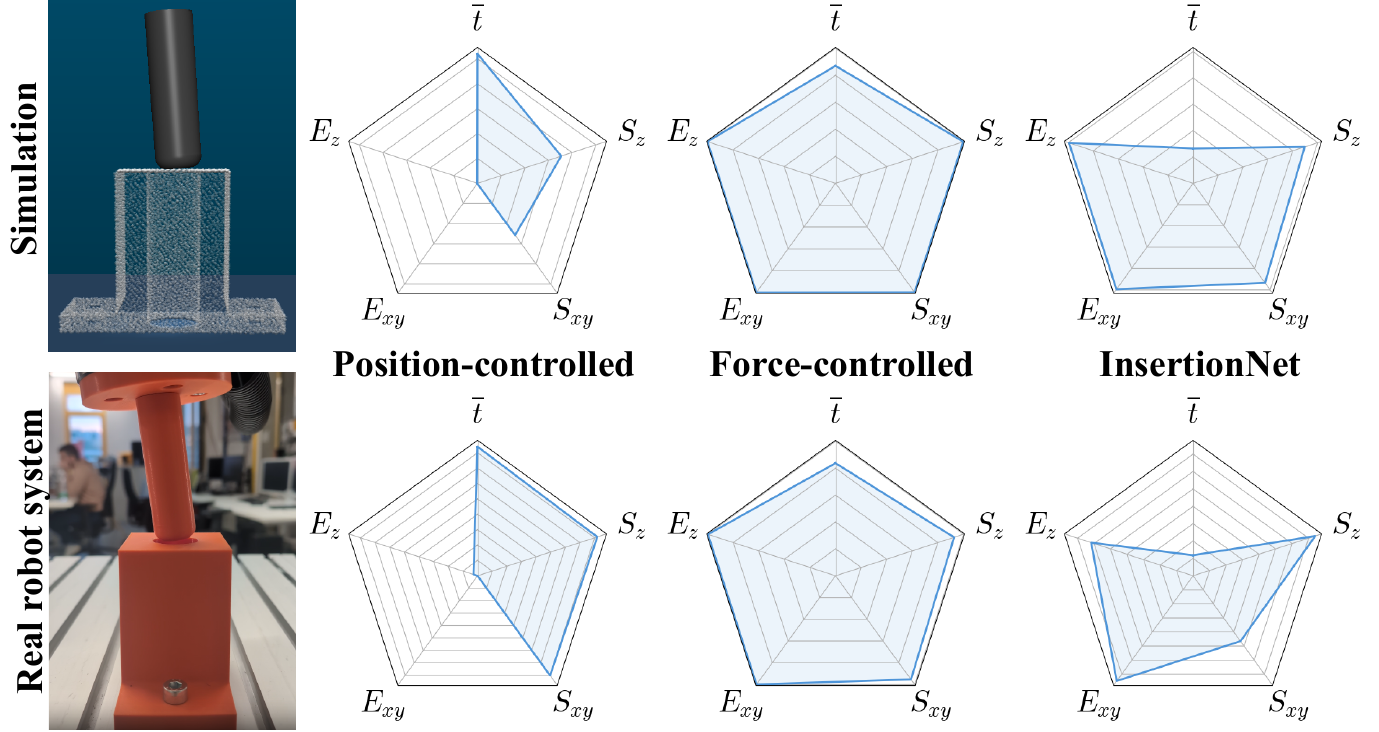}
     \caption{Resulting metrics for the insertion approaches in simulation (top row) and real robot system (bottom row). For each approach, 100 repetitions were conducted, using only successful trials.}
    \label{fig:comparison-metrics}
\end{figure}

\subsection{Insertion Metrics in Reality and Simulation}
On the real robotic setup, we sampled the uncertain initial pose $H_{s\Delta}$ in such a way, that task execution with each approach was successful.
We observed that InsertionNet performs more effectively in tasks with small insertion depths, which is reflected in the physical benchmark results shown in  Fig. \ref{fig:comparison-metrics}. 
The calculated metrics are normalized against the best result (with a slight offset), ensuring that the highest score for each metric is set to 1. 
Our proposed metrics effectively highlight differences between the various insertion approaches, as shown in Fig. \ref{fig:approach-comparison} and Fig. \ref{fig:comparison-metrics}. 

Position-controlled execution is susceptible to incurred forces and therefore yields lower scores for $E_z$ and $E_{xy}$, see Table \ref{tab:physical_benchmark_results}. As expected, the force-controlled approach performs best in minimizing the incurred forces in exchange for a longer completion time. Due to the use of force-torque data, InsertionNet is also able to solve the peg in hole task with comparable low incurred forces. However, as this is indirect and not direct control, the approach suffers from longer completion times. We replicated these experiments in simulation, observing a similar trend. However, discrepancies between real and simulated results can be attributed to differences in environmental properties. Especially Force Smoothness $S$ is dependent on accurate contact and friction modeling.

\begin{figure}
    \vspace{3mm}
    \centering
    \includegraphics[width=0.98\linewidth]{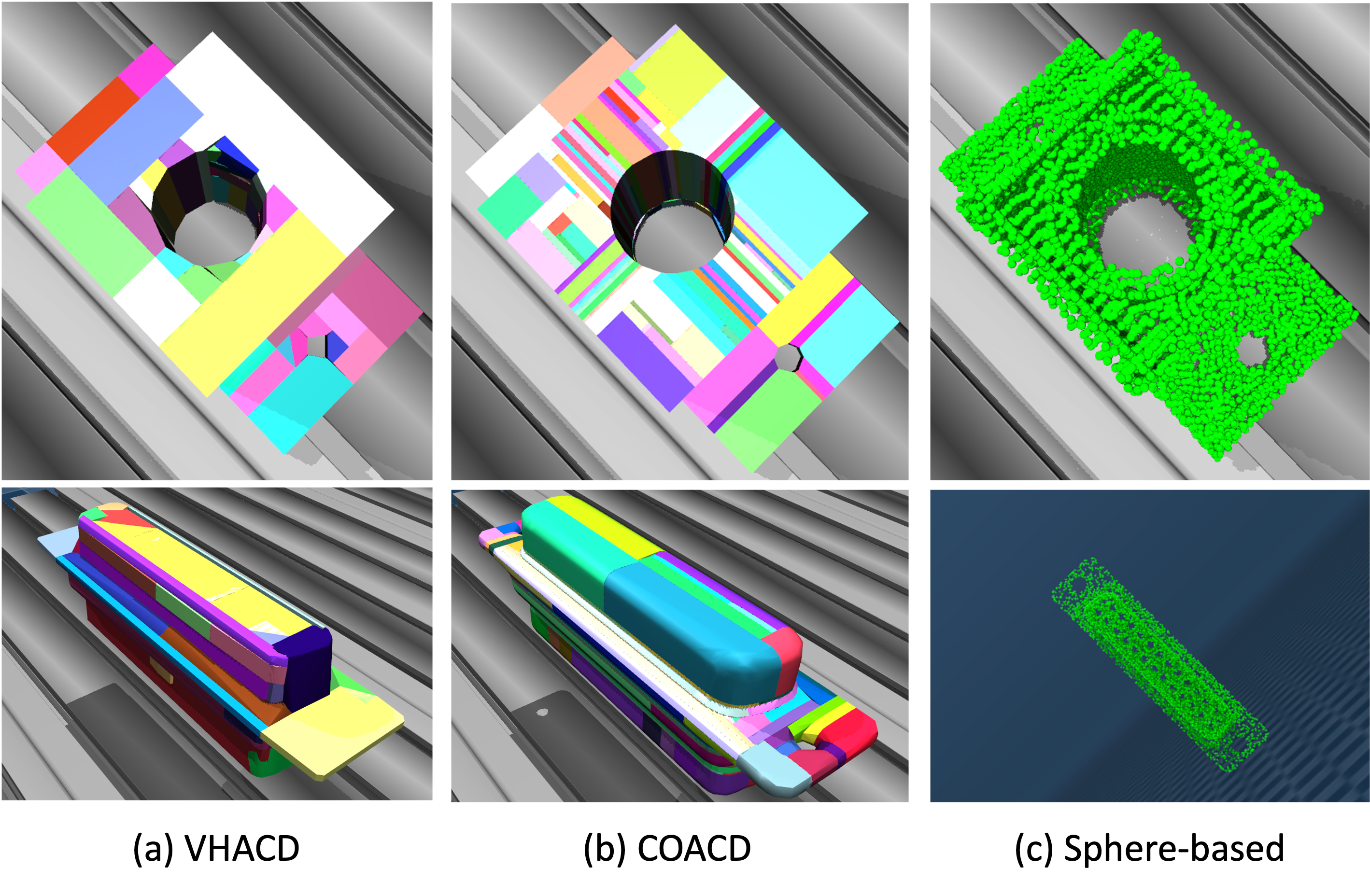}
    \caption{Mesh decomposition affects calculated contact forces in the simulation. Precise insertion task simulation requires high body contact density. Resulting decomposition of (a) VHACD, (b) COACD and (c) Sphere-based.}
    \label{fig:convex-decomposition}
\end{figure}
\begin{table}[!t]
\caption{Metrics of peg in hole task with physical robot}
\begin{tabular}{@{}l|ccccc@{}}
\toprule
baseline            & $\overline{t}$   & $E_z$ & $E_{xy}$ & $S_z$ & $S_{xy}$ \\ \midrule
Force-controlled    & 0.8375 & 0.9972 & 0.9999    & 0.9259 & 0.9534    \\
Position-controlled & 0.8395 & 0.0279 & 0.0002   & 0.8183 & 0.8016    \\
InsertionNet       & 0.1178 & 0.6173 & 0.7481    & 0.7406 & 0.4652    \\ \bottomrule
\end{tabular}
\label{tab:physical_benchmark_results}
\vspace{-2mm}
\end{table}

\begin{figure}
    \vspace{2mm}
    \centering
    \includegraphics[width=\linewidth]{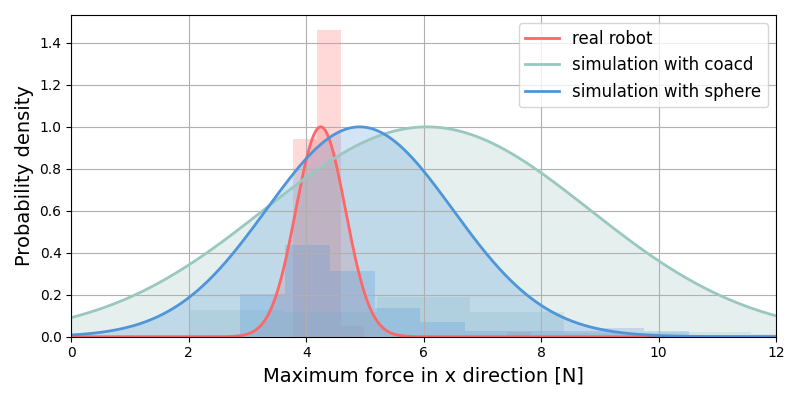}
    \caption{Maximum force distribution with randomized simulation parameters. Sphere based decomposition results in more realistic contact simulation and is less susceptible to parameter variation. }
    \label{fig:sim2real-gap}
\end{figure}

\subsection{Parameter Randomization in Simulation}
Given the complexity of contact modeling, we propose randomizing contact model parameters, including stiffness, impedance, and friction coefficients. This is achieved by adjusting global contact options such as \texttt{o\_solref}, \texttt{o\_solimp} and geometry \texttt{sliding friction} in MuJoCo. Since convex decomposition influences contact behavior in simulation, we apply different methods, namely VHACD, which is a standard method also used in PyBullet and COACD \cite{wei2022coacd}. Additionally, we propose a new sphere-based decomposition, see Fig. \ref{fig:convex-decomposition}.
Using sphere-based method, radius and quantity can be adjusted to increase the number of contacts. Additionally, spheres can be translated along their surface normals to simulate surface roughness, which is not possible with the other methods.

Experiments are conducted with 1 degree rotational error and executed 100 times on the real robot. For simulation we use COACD and the sphere-based method with randomized \texttt{o\_solref} in range of [0.01, 0.5], \texttt{o\_solimp} in range of [0.001, 0.99] and \texttt{sliding friction} varying from [0.1, 0.7]. We use 2000 spheres with a radius of 1 mm.

The objective is to leverage randomization to ensure the simulation-based large-scale test encompasses the distribution observed in the real world. Fig.~\ref{fig:sim2real-gap} illustrates that by randomizing the mentioned parameters, the simulation can effectively emulate the real-world scenario, thereby validating the efficacy of the algorithms in this broad-distributed contact condition. Consequently, it can be inferred that the algorithms will also perform satisfactorily in actual applications.

\subsection{Performance vs. Accuracy with simulated Friction}
\begin{figure}
    \centering
    \includegraphics[width=\linewidth]{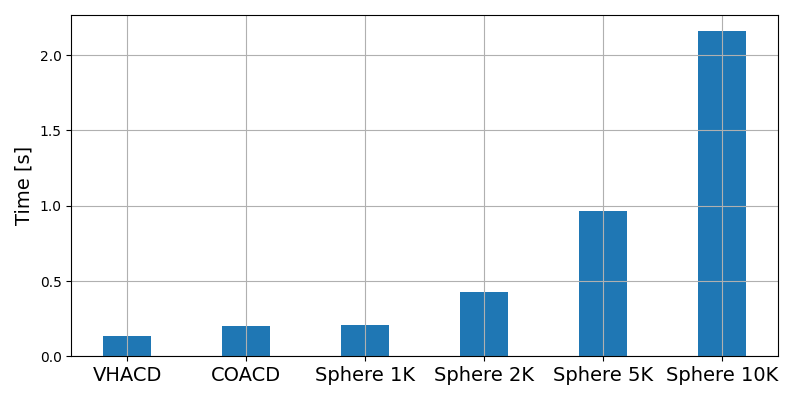}
    \caption{Time to run 1000 pure physics simulation steps with different mesh decompositions and mesh sizes.}
    \label{fig:time-cost-1k-steps}
    \vspace{-2mm}
\end{figure}
\begin{figure}
    \vspace{2mm}
    \centering
    \includegraphics[width=\linewidth]{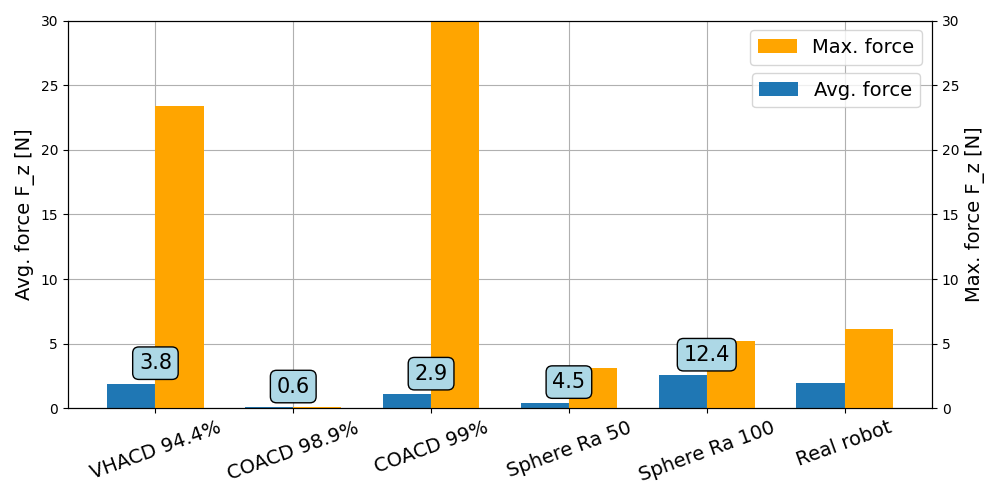}
    \caption{Maximum and average force during the insertion process and the average number of contacts used in the simulation to calculate the contact force.}
    \label{fig:force_in_insertion}
\end{figure}

The execution of precise insertion tasks necessitates the consideration of surface roughness.
To evaluate the simulation in this respect, we conduct real experiments with low tolerance of $0.1$ mm between peg and hole.
When 3D printing the peg, we randomize the seam position on the cylinder surface to achieve surface roughness.
Experiments are conducted with 0 positional error, $H_{s\Delta} = H_s$, and repeated 100 times on the real robot.
While the robot arm inserts the peg vertically into the hole, we measure the force in $z$-direction.
The same task is executed in simulation using different mesh decomposition approaches and compared with the results from the real experiments.
For the sphere-based method, we simulate the surface roughness using an arithmetic average roughness $R_a$ of $50 \mu$m and $100 \mu$m.
Fig. \ref{fig:force_in_insertion} shows that the sphere-based method yields similar average and maximum forces during insertion as real robot experiments. Since it uses more contact points per simulation step, it comes with the downside of increased simulation time, see Fig. \ref{fig:time-cost-1k-steps}. For VHACD and COACD, we minimally down-scaled the meshes to account for insertion friction. The minimal difference in scaling has a huge effect on the maximal force, as seen by down-scaling to 98.9\% and 99\% for COACD, which makes current decomposition methods impractical to use.

\subsection{Cloud-based Scaling}

\begin{table}[t]
\begin{center}
    \vspace{-0.5mm}
	\caption{1000 Runs: Simulation Execution Performance}
    \label{tab:cloud_scale}
	\begin{tabular}{cccc}
		\Xhline{2\arrayrulewidth}
		\\[-0.7em]
		& \multirow{2}{*}{\begin{tabular}[c]{@{}c@{}}Sequential\\execution\end{tabular}}& \multicolumn{2}{c}{Cloud-Accelerated}  \\[0.1em] \cline{3-4}  
		\\[-0.7em]
		Experiment       &           & \multicolumn{1}{c}{\begin{tabular}[c]{@{}c@{}} 1 nodes\end{tabular}}  & \multicolumn{1}{c}{\begin{tabular}[c]{@{}c@{}} 3 nodes \end{tabular}} \\ \Xhline{1\arrayrulewidth}
		\\[-0.5em]
		\begin{tabular}[c]{@{}c@{}}Small Net + Simple Sim\end{tabular}   &  2.8 h   &  0.51 h   &   0.18 h \\
		\\[-0.7em]
		\begin{tabular}[c]{@{}c@{}}Small Net + Precise Sim\end{tabular}  &  6.95 h   &   0.93 h  &   0.33 h \\ 
        \\[-0.7em]
        \begin{tabular}[c]{@{}c@{}}Large Net + Simple Sim\end{tabular}  &  14.9 h   &   1.83 h  &   0.69 h \\ 
        \\[-0.7em]
        \begin{tabular}[c]{@{}c@{}}Large Net + Precise Sim\end{tabular}  & 16.3 h   &  2.34 h  &   0.82 h \\ 
		\Xhline{2\arrayrulewidth}
		\\[-0.7em]
		\multicolumn{4}{l}{\footnotesize Equipped with AMD Ryzen 7 5700X and Nvidia RTX 4090} \\
	\end{tabular}
\vspace{-6mm}
\end{center}
\end{table}

The combination of parameter randomization in the simulation dramatically increased the uncertainty in the perception of the starting pose, the number of task objects tested, and the number of test runs. To evaluate the necessity and effectiveness of using the cloud-based scaling approach, we conducted the experiment with InsertionNet trained with different backbones, one with the Yolo-lite as a small network and another with ResNet50 as a large network. Both networks are implemented in TensorFlow. In the simulation, we use two different setups, one with 0.1 ms as time step, which runs fast and gives acceptable results, and another with 1 ms as time step, which gives better accuracy but takes the most time to run the physical simulation steps.

We measure the time cost of running 1000 insertion processes on three different setups, first, the classical approach that runs in sequence, the second and third are on the cluster to speed up. We use asymmetric scaling of the inference and simulation containers, since the inference and simulation time are different. The results in Tab.~\ref{tab:cloud_scale} show the acceleration with the parallelized execution.
With Kubernetes-based infrastructure, it is easy to add more resources for further acceleration. MuJoCo's physical engine is also well-optimized for CPU execution. Since numerical integration is serial, using more CPU-optimized computing instances can increase the GPU utilization of the cluster even more. In summary, the use of cloud computing is necessary to enable this comprehensive evaluation with simulation.

\section{CONCLUSION AND OUTLOOK}
\label{sec:conclusion}

In this paper, we introduced QBIT, a cloud-based benchmarking framework designed for comprehensive evaluation of robotic insertion tasks. By incorporating quality-aware metrics beyond success rate, our framework provides a more detailed assessment of insertion performance while accounting for perception uncertainty.

QBIT is designed in a fully microservice-oriented architecture and provides interfaces to both simulation and real robots.
Through comparative analysis of three different insertion approaches, we demonstrated that our proposed metrics effectively distinguish qualitative differences in insertion performance, even when all executions were technically successful.

Furthermore, we proposed to use a sphere-based decomposition approach in the simulation to achieve a near-realistic contact behavior, since the convex decomposition approach is extremely sensitive to mesh scale.
Regarding the sim-to-real gap problem, we proposed to randomize the parameters in the contact models to increase the possibility that the distribution in the real world is covered by the simulation.

To manage the large-scale simulations required for parameter randomization and uncertainty modeling, we leveraged cloud computing and Kubernetes-based parallelization.  With its modular and extensible design, QBIT serves as a valuable tool for robotic insertion research, facilitating the transition of novel approaches from lab-scale experimentation to real-world industrial applications.

Although our approach showed promising results, several areas remain open for future improvements.
First, expanding the framework to support a broader range of robotic platforms is essential. Currently, we have integrated robots available in our lab, but we provide a template class to facilitate the inclusion of additional robots. Second, given the strong performance of MuJoCo in simulating contact interactions, our framework can be extended to other contact-rich manipulation tasks that require high-fidelity physical simulations. Lastly, we found that the built-in rendering is limited and can be improved by using an external rendering library.

\section*{ACKNOWLEDGMENT}
Part of this research is being conducted as part of the KI5GRob project funded by the German Federal Ministry of Education and Research (BMBF) under project number 13FH579KX9. 


\bibliographystyle{IEEEtran} 
\bibliography{IEEEabrv, refs}

\begin{thebibliography}{10}
\providecommand{\url}[1]{#1}
\csname url@rmstyle\endcsname
\providecommand{\newblock}{\relax}
\providecommand{\bibinfo}[2]{#2}
\providecommand\BIBentrySTDinterwordspacing{\spaceskip=0pt\relax}
\providecommand\BIBentryALTinterwordstretchfactor{4}
\providecommand\BIBentryALTinterwordspacing{\spaceskip=\fontdimen2\font plus
\BIBentryALTinterwordstretchfactor\fontdimen3\font minus
  \fontdimen4\font\relax}
\providecommand\BIBforeignlanguage[2]{{%
\expandafter\ifx\csname l@#1\endcsname\relax
\typeout{** WARNING: IEEEtran.bst: No hyphenation pattern has been}%
\typeout{** loaded for the language `#1'. Using the pattern for}%
\typeout{** the default language instead.}%
\else
\language=\csname l@#1\endcsname
\fi
#2}}

\bibitem{Müller_Kutzbach_2019}
C.~Müller and N.~Kutzbach, ``World robotics 2019 – industrial robots,''
  2019.

\bibitem{Whitney_2004}
\emph{\BIBforeignlanguage{eng}{Mechanical assemblies: their design,
  manufacture, and role in product development}}, ser. Oxford series on
  advanced manufacturing.\hskip 1em plus 0.5em minus 0.4em\relax New York:
  Oxford University Press, 2004.

\bibitem{Zhang_Tomizuka_Li_2024}
\BIBentryALTinterwordspacing
X.~Zhang, M.~Tomizuka, and H.~Li, ``Bridging the sim-to-real gap with dynamic
  compliance tuning for industrial insertion,'' in \emph{2024 IEEE
  International Conference on Robotics and Automation (ICRA)}.\hskip 1em plus
  0.5em minus 0.4em\relax Yokohama, Japan: IEEE, May 2024, p. 4356–4363.
  [Online]. Available: \url{https://ieeexplore.ieee.org/document/10610707/}
\BIBentrySTDinterwordspacing

\bibitem{Marvel_Falco_2012}
\BIBentryALTinterwordspacing
J.~Marvel and J.~Falco, \emph{\BIBforeignlanguage{en}{Best Practices and
  Performance Metrics Using Force Control for Robotic Assembly}}, 0th~ed.,
  Gaithersburg, MD, Nov. 2012, no. NIST IR 7901. [Online]. Available:
  \url{https://nvlpubs.nist.gov/nistpubs/ir/2012/NIST.IR.7901.pdf}
\BIBentrySTDinterwordspacing

\bibitem{Sliwowski_Jadav_Stanovcic_Orbik_Heidersberger_Lee_2025}
\BIBentryALTinterwordspacing
D.~Sliwowski, S.~Jadav, S.~Stanovcic, J.~Orbik, J.~Heidersberger, and D.~Lee,
  ``Reassemble: A multimodal dataset for contact-rich robotic assembly and
  disassembly,'' no. arXiv:2502.05086, Feb. 2025, arXiv:2502.05086 [cs].
  [Online]. Available: \url{http://arxiv.org/abs/2502.05086}
\BIBentrySTDinterwordspacing

\bibitem{james2020rlbench}
S.~James, Z.~Ma, D.~R. Arrojo, and A.~J. Davison, ``Rlbench: The robot learning
  benchmark {\&} learning environment,'' \emph{IEEE Robotics and Automation
  Letters}, vol.~5, no.~2, pp. 3019--3026, 2020.

\bibitem{Luo_Li_2021}
\BIBentryALTinterwordspacing
J.~Luo and H.~Li, ``A learning approach to robot-agnostic force-guided high
  precision assembly,'' in \emph{2021 IEEE/RSJ International Conference on
  Intelligent Robots and Systems (IROS)}.\hskip 1em plus 0.5em minus
  0.4em\relax Prague, Czech Republic: IEEE, Sept. 2021, p. 2151–2157.
  [Online]. Available: \url{https://ieeexplore.ieee.org/document/9636328/}
\BIBentrySTDinterwordspacing

\bibitem{Zang_Wang_Pan_Hou_Ding_Zhao_2024}
\BIBentryALTinterwordspacing
Y.~Zang, Z.~Wang, M.~Pan, Z.~Hou, Z.~Ding, and M.~Zhao, ``Safe peg-in-hole
  automatic assembly using virtual guiding force: A deep reinforcement learning
  solution,'' 2024. [Online]. Available:
  \url{https://www.ssrn.com/abstract=4862365}
\BIBentrySTDinterwordspacing

\bibitem{Shi_Yuan_Tsitos_Cong_Hadjar_Chen_Zhang_2023}
Y.~Shi, C.~Yuan, A.~Tsitos, L.~Cong, H.~Hadjar, Z.~Chen, and J.~Zhang, ``A
  sim-to-real learning-based framework for contact-rich assembly by utilizing
  cyclegan and force control,'' \emph{IEEE Transactions on Cognitive and
  Developmental Systems}, vol.~15, no.~4, p. 2144–2155, Dec. 2023.

\bibitem{Wang_Beltran-Hernandez_Wan_Harada_2021}
\BIBentryALTinterwordspacing
Y.~Wang, C.~C. Beltran-Hernandez, W.~Wan, and K.~Harada, ``Robotic imitation of
  human assembly skills using hybrid trajectory and force learning,'' in
  \emph{2021 IEEE International Conference on Robotics and Automation
  (ICRA)}.\hskip 1em plus 0.5em minus 0.4em\relax Xi’an, China: IEEE, May
  2021, p. 11278–11284. [Online]. Available:
  \url{https://ieeexplore.ieee.org/document/9561619/}
\BIBentrySTDinterwordspacing

\bibitem{Ankile_Simeonov_Shenfeld_Agrawal_2024}
\BIBentryALTinterwordspacing
L.~Ankile, A.~Simeonov, I.~Shenfeld, and P.~Agrawal, ``Juicer: Data-efficient
  imitation learning for robotic assembly,'' 2024. [Online]. Available:
  \url{https://arxiv.org/abs/2404.03729}
\BIBentrySTDinterwordspacing

\bibitem{Wang_Su_Sun_Chen_Xie_2024}
C.~Wang, C.~Su, B.~Sun, G.~Chen, and L.~Xie, ``Extended residual learning with
  one-shot imitation learning for robotic assembly in semi-structured
  environment,'' \emph{Frontiers in Neurorobotics}, vol.~18, p. 1355170, Apr.
  2024.

\bibitem{Wang_Liu_Liu_Huang_Yang_2024}
G.~Wang, X.~Liu, Z.~Liu, P.~Huang, and Y.~Yang, ``Visual-tactile perception
  based control strategy for complex robot peg-in-hole process via topological
  and geometric reasoning,'' \emph{IEEE Robotics and Automation Letters},
  vol.~9, no.~10, p. 8410–8417, Oct. 2024.

\bibitem{Mei_Li_Tan_Zhu_Li_Zhou_2024}
Y.~Mei, R.~Li, Y.~Tan, D.~Zhu, T.~Li, and Z.~Zhou, ``Robotic assembly strategy
  with wrist force sense for narrow clearance peg-in-hole,'' \emph{IEEE
  Transactions on Automation Science and Engineering}, p. 1–16, 2024.

\bibitem{Bielak_2024}
\BIBentryALTinterwordspacing
J.~Bielak, \emph{\BIBforeignlanguage{en}{The Finite Element Method: A
  Primer}}.\hskip 1em plus 0.5em minus 0.4em\relax Cham: Springer International
  Publishing, 2024. [Online]. Available:
  \url{https://link.springer.com/10.1007/978-3-031-56369-0}
\BIBentrySTDinterwordspacing

\bibitem{coumans2020}
E.~Coumans and Y.~Bai, ``Pybullet, a python module for physics simulation for
  games, robotics and machine learning,'' \url{http://pybullet.org},
  2016--2023.

\bibitem{koenig2004gazebo}
N.~Koenig and A.~Howard, ``Design and use paradigms for gazebo, an open-source
  multi-robot simulator,'' in \emph{2004 IEEE/RSJ international conference on
  intelligent robots and systems (IROS)(IEEE Cat. No. 04CH37566)},
  vol.~3.\hskip 1em plus 0.5em minus 0.4em\relax IEEE, 2004, pp. 2149--2154.

\bibitem{todorov2012mujoco}
E.~Todorov, T.~Erez, and Y.~Tassa, ``Mujoco: A physics engine for model-based
  control,'' in \emph{2012 IEEE/RSJ international conference on intelligent
  robots and systems}.\hskip 1em plus 0.5em minus 0.4em\relax IEEE, 2012, pp.
  5026--5033.

\bibitem{erez2015simulation}
T.~Erez, Y.~Tassa, and E.~Todorov, ``Simulation tools for model-based robotics:
  Comparison of bullet, havok, mujoco, ode and physx,'' in \emph{2015 IEEE
  international conference on robotics and automation (ICRA)}.\hskip 1em plus
  0.5em minus 0.4em\relax IEEE, 2015, pp. 4397--4404.

\bibitem{acosta2022validating}
B.~Acosta, W.~Yang, and M.~Posa, ``Validating robotics simulators on real-world
  impacts,'' \emph{IEEE Robotics and Automation Letters}, vol.~7, no.~3, pp.
  6471--6478, 2022.

\bibitem{todorov2011contact}
E.~Todorov, ``A convex, smooth and invertible contact model for trajectory
  optimization,'' in \emph{2011 IEEE International Conference on Robotics and
  Automation}.\hskip 1em plus 0.5em minus 0.4em\relax IEEE, 2011, pp.
  1071--1076.

\bibitem{todorov2014contact}
{Todorov, Emanuel}, ``Convex and analytically-invertible dynamics with contacts
  and constraints: Theory and implementation in mujoco,'' in \emph{2014 IEEE
  International Conference on Robotics and Automation (ICRA)}.\hskip 1em plus
  0.5em minus 0.4em\relax IEEE, 2014, pp. 6054--6061.

\bibitem{liu2021ocrtoc}
Z.~Liu, W.~Liu, Y.~Qin, F.~Xiang, M.~Gou, S.~Xin, M.~A. Roa, B.~Calli, H.~Su,
  Y.~Sun, \emph{et~al.}, ``Ocrtoc: A cloud-based competition and benchmark for
  robotic grasping and manipulation,'' \emph{IEEE Robotics and Automation
  Letters}, vol.~7, no.~1, pp. 486--493, 2021.

\bibitem{aws2025robomaker}
{Amazon Web Services, Inc.}, ``Aws robomaker,''
  https://aws.amazon.com/robomaker, Accessed: 2025-02-28.

\bibitem{lin2023ems}
Q.~Lin, G.~Ye, and H.~Liu, ``Ems{\textregistered}: A massive computational
  experiment management system towards data-driven robotics,'' in \emph{2023
  IEEE International Conference on Robotics and Automation (ICRA)}.\hskip 1em
  plus 0.5em minus 0.4em\relax IEEE, 2023, pp. 9068--9075.

\bibitem{ichnowski2023fogros2}
J.~Ichnowski, K.~Chen, K.~Dharmarajan, S.~Adebola, M.~Danielczuk,
  V.~Mayoral-Vilches, H.~Zhan, D.~Xu, R.~Ghassemi, J.~Kubiatowicz,
  \emph{et~al.}, ``Fogros 2: An adaptive and extensible platform for cloud and
  fog robotics using ros 2,'' in \emph{Proceedings IEEE International
  Conference on Robotics and Automation}, 2023.

\bibitem{zhang2024conquering}
Y.~Zhang, F.~Pasch, F.~Mirus, K.-U. Scholl, C.~Wurll, and B.~Hein, ``Conquering
  the robotic software development cycle at scale: Using kuberos from
  simulation to real-world deployment,'' in \emph{2024 IEEE 20th International
  Conference on Automation Science and Engineering (CASE)}.\hskip 1em plus
  0.5em minus 0.4em\relax IEEE, 2024, pp. 1503--1509.

\bibitem{zhang2023kuberos}
Y.~Zhang, C.~Wurll, and B.~Hein, ``Kuberos: A unified platform for automated
  and scalable deployment of ros2-based multi-robot applications,'' in
  \emph{2023 IEEE International Conference on Robotics and Automation
  (ICRA)}.\hskip 1em plus 0.5em minus 0.4em\relax IEEE, 2023, pp. 9097--9103.

\bibitem{grpc2025}
{grpc.io}, ``grpc,'' https://grpc.io, Accessed: 2025-02-28.

\bibitem{macenski2022robot}
S.~Macenski, T.~Foote, B.~Gerkey, C.~Lalancette, and W.~Woodall, ``Robot
  operating system 2: Design, architecture, and uses in the wild,''
  \emph{Science Robotics}, vol.~7, no.~66, 2022.

\bibitem{Coleman2014ReducingTB}
D.~Coleman, I.~A. Sucan, S.~Chitta, and N.~Correll, ``Reducing the barrier to
  entry of complex robotic software: a moveit! case study,'' \emph{Journal of
  Software Engineering for Robotics}, 2014.

\bibitem{yoshikawa2000force}
T.~Yoshikawa, ``Force control of robot manipulators,'' in \emph{Proceedings
  2000 ICRA. Millennium Conference. IEEE International Conference on Robotics
  and Automation. Symposia Proceedings (Cat. No. 00CH37065)}, vol.~1.\hskip 1em
  plus 0.5em minus 0.4em\relax IEEE, 2000, pp. 220--226.

\bibitem{Spector_Castro_2021}
O.~Spector and D.~D. Castro, ``Insertionnet - a scalable solution for
  insertion,'' \emph{IEEE Robotics and Automation Letters}, vol.~6, no.~3, p.
  5509–5516, July 2021.

\bibitem{wei2022coacd}
X.~Wei, M.~Liu, Z.~Ling, and H.~Su, ``Approximate convex decomposition for 3d
  meshes with collision-aware concavity and tree search,'' \emph{ACM
  Transactions on Graphics (TOG)}, vol.~41, no.~4, pp. 1--18, 2022.

\end{thebibliography}

\end{document}